\documentclass[letterpaper, 10 pt, compsocconf]{IEEEtran}  

\IEEEoverridecommandlockouts                              

\usepackage{comment}

\usepackage{cite}
\usepackage{amsmath,amssymb,amsfonts}
\usepackage{graphicx}
\usepackage{multicol}
\usepackage{textcomp}
\usepackage{xcolor}
\usepackage{caption}
\usepackage{subfigure}
\usepackage{makecell}

\usepackage{cite}
\usepackage{hyperref}
\usepackage{url}
\hypersetup{colorlinks=true,
linkcolor=green,
hypertexnames=false,
pdfhighlight=/N
}

\usepackage{amssymb}
\usepackage{array}
\usepackage{tabularx}

\usepackage{graphics} 
\usepackage{amsmath} 

\usepackage{commath}

\usepackage{algorithm}
\usepackage[noend]{algpseudocode}

\usepackage{caption}
\usepackage{subfigure}

\usepackage{multirow}
\usepackage{setspace}
\usepackage{caption}

\usepackage{tikz}

\usepackage{multirow}

\usepackage{float}
\usepackage{wrapfig}

\usepackage{dblfloatfix}    
\usepackage{kantlipsum}     

\title{\LARGE \bf
Deep Learning with Attention Mechanism for Predicting Driver Intention at Intersection
}

\author{Abenezer Girma$^{1}$, Seifemichael Amsalu$^{2}$, Abrham Workineh$^{3}$, Mubbashar Khan$^{1}$ and Abdollah Homaifar$^{1*}$

\thanks{*Corresponding Author: A. Homaifar, Telephone: (336) 2853271}
\thanks{$^{1}$Abenezer Girma, Mubbashar Khan and Abdollah Homaifar are with Faculty of Electrical and Computer Engineering and the Autonomous Control of Information Technology Institute, North Carolina A$\&$T State University, Greensboro, NC 27411 USA {\tt\small @aggies.ncat.edu, makhan1@ncat.edu ,homaifar@ncat.edu}}%

\thanks{$^{2}$Seifemichael Amsalu is a Prototype Engineer at Intel Corporation, Portland, Oregon,
USA  {\tt\small seifemichael.amsalu@intel.com}}%

\thanks{$^{2}$Abrham Workineh is a Senior Data Scientist at USAA, San Antonio, Texas, USA {\tt\small abrham.workineh@usaa.com}}%
}

\begin{document}

\maketitle

\begin{abstract}
In this paper, a driver's intention prediction near a road intersection is proposed. Our approach uses a deep bidirectional Long Short-Term Memory (LSTM) with an attention mechanism model based on a hybrid-state system (HSS) framework. As intersection is considered to be as one of the major source of road accidents, predicting a driver's intention at an intersection is very crucial. Our method uses a sequence to sequence modeling with an attention mechanism to effectively exploit temporal information out of the time-series vehicular data including velocity and yaw-rate. The model then predicts ahead of time whether the target vehicle/driver will go straight, stop, or take right or left turn. The performance of the proposed approach is evaluated on a naturalistic driving dataset and results show that our method achieves high accuracy as well as outperforms other methods. The proposed solution is promising to be applied in advanced driver assistance systems (ADAS) and as part of active safety system of autonomous vehicles.        

\end{abstract}


%
\IEEEpeerreviewmaketitle

\section{INTRODUCTION}
In the past few years, more than 50$\%$ of the combined injury and fatal accidents happen at or near intersection \footnote{https://highways.dot.gov/research/research-programs/safety/intersection-safety}. Recent autonomous vehicles development results in a diverse-traffic environment where human-driven and autonomous vehicles are suppose to share the same road. 
To ensure safe and sustained traffic flow in such environment, addressing root-source of intersection accident is very important. Accordingly, effectively estimating and predicting the drivers' intention at intersection is very crucial to address the cause of accident and assure road safety in such a heterogeneous environment. In automated driving, Advanced Driver Assistance Systems (ADAS) also requires such a technique of driver's intention estimation \cite{jones2001keeping} for ensuring safe driving. 

In this paper, a multi-agent driver behavior model is proposed that estimates the drivers' intention at intersection. The model uses a framework developed in \cite{kurt2011probabilistic,song2000human } machine learning model, which uses different mathematical or symbolic methods to simulates the attention/cognition and control behavior of interest, based on dynamic environmental data \cite{song2000human, ozguner2011autonomous}.For example, in road intersection near-crash activity,  the goal is to use a set of observations to be able to estimate whether the driver is keep going stright, stop, turn left, turn right safely in accordance with traffic signal indicator.  As different drivers have different driving behavior, the differences in observation must be considered in the prediction process. For instance, at an intersection, when the driver decides to "Turn Left," there are sequences of actions that need to be carried out, the turn left signal blinks, brake light illuminate, the car slows down, and the vehicle turns left finally. Based on the observation of such continuous vehicle dynamics, the driver's behavior estimation attempts to estimate the driver's decisions that result those observed vehicle dynamics.  The term "driver" is used to refer to the vehicle and the driver together,  whereas "driver behavior" referred to the trajectory shown by the driver \cite{gadepally2013framework}. 
In our study, the potential of deep learning model with attention mechanism to predict the drivers' intention from a time series naturalistic driving data that represents the continuous vehicle dynamics is investigated for driver intention prediction at road intersection. Hybrid-State System (HSS) framework are used as a base framework for this implementation\cite{kurt2010hybrid}.  

Many researchers have studied the modeling of driver behaviors. In \cite{oliver2000graphical}, the authors developed graphical models and Hidden Markov Models (HMMs) focusing effect of context on the driver's performance. In addition to real-time vehicular data, the surrounding contextual information is used as an input to estimate the maneuvers. In \cite{kuge2000driver}, driver behavior identification for cases of emergency and typical lane changes task is examined for a human behavior cognitive model  using an HMM method. A driving simulator generated data is used to train the HMM models. In \cite{mitrovic2005reliable}, a model for identification of driving events using Discrete HMMs (DHMMs) is proposed that uses acceleration and velocity data from a real vehicle recroded in a normal driving environment. In \cite{amsalu2015driver}, Support Vector Machines (SVM) and Gaussian Mixture Model (GMM) in combination with HMM and Gaussian Mixture Model (GMM) are used to predict drivers intention near intersection using a naturalistic data. The work has extended by incorporating genetic algorithm with HMM to improve the identification performance of the model \cite{amsalu2016driver,  amsalu2017driver}. 

All the proposed techniques fall short as the implemented machine learning models can not effectively capture long term temporal relationship in the data to predict a near-crash event ahead of time. However, long-short-term-memory (LSTM) type of deep learning models have been proven efficient in modeling sequential tasks like the time series vehicular data, because it stores the memory of the context in the sequence. Over the last few years and with the recent advances in the field of deep learning, researchers have utilized variant of these methods to solve driver behavior modeling tasks \cite{zyner2018recurrent, girma2019driver}. In \cite{dang2017time}, the time to lane change of a vehicle is estimated using LSTM combined with a regression approach. In \cite{zhao2017speed}, Deep Belief Network (DBN) is used to learn and predict speed and steering angle from naturalistic driving data. Driver behavior analysis of vehicles surrounding the ego-vehicle is conducted in \cite{zyner2018recurrent} using recurrent neural network model.

In this paper, we present a solution using deep bidirectional LSTM with attention mechanism based on HSS framework to predict driver intention at intersection ahead of time. We examined the driver intention estimation as a sequence to sequence modeling, meaning the model aims to map a sequence of sensor observation to a sequence of estimated driver intention. The bidirectional part of the LSTM model captures long term dependencies in the sequence and the attention mechanism enables the model to learn where to give focus in the sequence to increase the overall model performance.

The paper is organized as follows: section II presents the necessary strategies and resources used to solve the driver behavior modeling problem that includes explanation of the HSS framework and the bidirectional deep LSTM with attnention mechanism model.
Section III discuss the data collection and Analysis procedures and Section IV presents the results of the proposed approach. Finally, the conclusion and future work are presented in Section V. 

\section{DRIVER INTENTION PREDICTION FRAMEWORK}
\begin{figure}[thpb]
   \centering
    \includegraphics[width=0.4\textwidth]{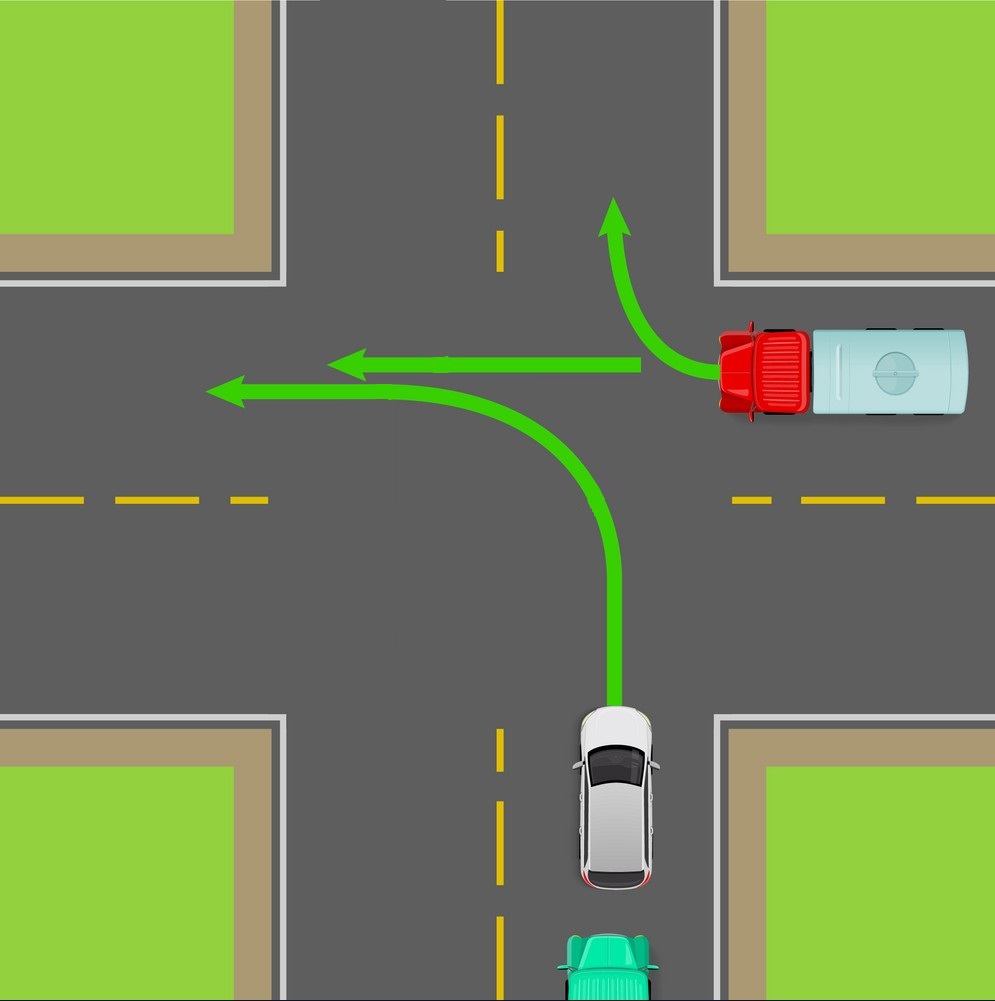}
    \captionsetup{format=hang}
    \caption{
Example near intersection driving. The red car has the right of way. Before turning left the silver car must first determine the intention of the red car using the proposed model.}
    \label{fig:intersection}
\end{figure}

In driver behavior modeling, the estimation of the driver's intention is done from the vehicle dynamics, as explained above. Figure \ref{fig:intersection} shows an example of near intersection driving as the silver car (host) approaches the road intersection and tends to turn left. This car must determine the intention of the red vehicle (target) that has the right of way. If the target vehicle turns right, the host can turn left. If not, the host car must stop and determine the intention of the target car driver using the proposed model from the driving data obtained through vehicle-to-vehicle (V2V) communication or through the use of on-board sensors such as lidar and radar\cite{gadepally2013framework}.

Likewise, the intersection scenario explained above, the driver behavior modeling framework for other driving situations of interest, such as lane change, and other near-crash events, can be formulated.
\subsection{Hybrid State System (HSS) Framework}

\begin{figure}[thpb]
   \centering
    \includegraphics[width=0.3\textwidth]{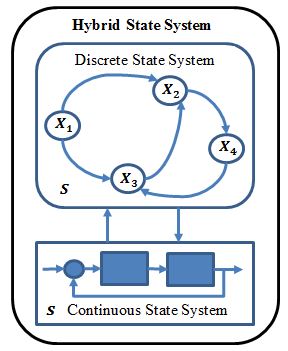}
    \caption{Hybrid State System (HSS) model.}
    \label{fig:hybridstate}
\end{figure}

In different applications, that includes autonomous vehicles, Hybrid-State System (HSS) widely used to model hierarchically the interaction of a discrete-state system and a continuous-state \cite{kurt2010hybrid, amsalu2017driver}. The HSS model can be used to capture the interaction of the vehicle and its driver and also applied to estimate and predict their behaviors.  Figure \ref{fig:hybridstate} shows the HSS setting containing the discrete-state system (DSS) at a higher level and a continuous-state system (CSS) in the lower level. HSS represents the interaction of the driver and the vehicle. The driver responds based on non-continuous(discrete) events and gives corresponding higher level driving decisions that affect the lower level continuous vehicle dynamics/trajectory.   The HSS system formulation is developed in \cite{amsalu2017driver}.

The DSS and CSS interaction that exist in the HSS framework can be mathematically modeled using machine learning approach. In this work, deep LSTM is used for this modeling \cite{zhao2017speed} to effectively capture the stochastic relationship between the driver's state and dynamics of the vehicle. Here, the changes in driver states are predicted from the changes happening in vehicle dynamics using the naturalistic driving data based on the proposed model.

\subsection{Deep LSTM}
In this section, the summary of LSTM types of RNN architecture is presented. In recent years LSTM have been successful in time series prediction tasks such as language translation, driver identification and human activity recognition \cite{girma2019driver, lara2012survey}. Modeling sequential tasks using LSTMs is effective, because internal artificial memory of LSTMs' can easily capture the temporal relationships exist in time-series data. 
The internal states of LSTM are called "memory blocks" that are regulated by three structures called "gates", which are: forget gate, input gate and output gate. These gates work together in order to create a memory in the network that captures the temporal dependencies in the data.



In unidirectional LSTM, information is passed only from the past to the future inputs \cite{gers1999learning}. Bidirectional LSTM is the extension of unidirectional LSTM where information flows to and from both the past and the future time steps \cite{graves2005framewise}. As a result bidirectional LSTM performs better than the unidirectional. Furthermore, ordinary LSTM memory blocks attempts to learn single vector representation, however attention mechanism helps LSTM to learn how to ‘attend to’ (focus on) to most important input sequence vectors based on learnable attention weights \cite{bahdanau2014neural, vaswani2017attention}.

\begin{figure}[!h]
   \centering
    \includegraphics[width=0.46\textwidth]{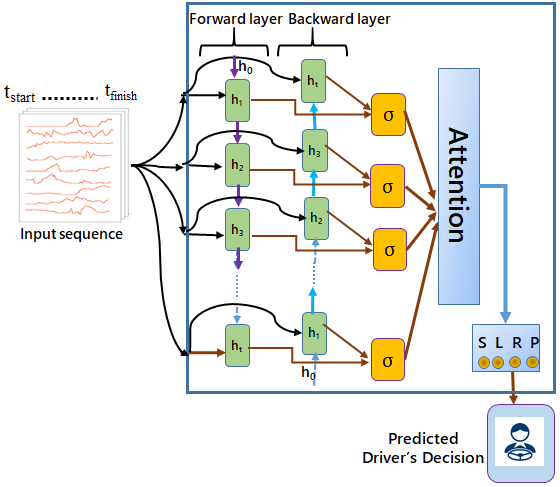}
    \captionsetup{format=hang}
    \caption{Bidirectional LSTM with attention mechanism: takes input sequence and predict driver's decision as S(Straight), L(left), R(right) and P(Stop).}    
    \label{fig:bilstm}
\end{figure}

In this paper, as shown in Fig \ref{fig:bilstm}, we combined a bidirectional LSTM with attention mechanism to build a \textit{\textbf{bidirectional LSTM with an attention model}} to extract the most important information out of the time-series vehicular data. The overall problem is formulated as a Neural Machine Translation, sequence to sequence model \cite{khosroshahi2016surround}.

\begin{figure}[thpb]
   \centering
\includegraphics[width=0.5\textwidth]{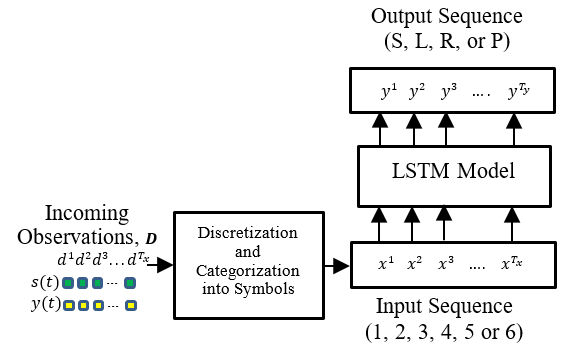}
    \captionsetup{format=hang}
    \caption{ The architecture of the proposed technique. It estimates the possible driver's intention from new observations sequence using LSTM as one of the intentions which are  Right (R), Left (L), Straight (S) and Stop (P).}
    \label{fig: arch}
\end{figure}

In Figure \ref{fig: arch}, the overall architecture of the proposed technique is shown. The model takes sequence of input symbols at each time steps and predicts the driver's intention near intersection.  The input symbols are the discretized and categorized in form of the continuous vehicle maneuver data of velocity and yaw-rate, which is discussed in detail in the next section. The possible driver action at the intersection is represented as S, L, R, and P for Straight, Left, Right, and Stop driver maneuvers, respectively. The implementation of the proposed model is done using Python \footnote{https://www.python.org/} programming language and with Keras \footnote{https://keras.io/} deep learning framework.  

\section{DATA COLLECTION AND ANALYSIS }
Naturalistic Driving Study (NDS) has been used in order to show the role of driver behavior modeling and to evaluate its performance in increasing traffic safety.
The NDS dataset collected by the Ohio State University \cite{gadepally2013framework} is used for training and evaluating the proposed approach. 
Honda Accord 2012 model vehicle has been used as a target vehicle to collect the data. We assumed the host vehicle obtains the target vehicle data through vehicle-to-vehicle communication \cite{gadepally2013framework, ashraf2017towards} to estimate the target vehicle intention. Low-level continuous observations data including yaw-rate, steering wheel angle, velocity, odometer, and acceleration as well as video and GPS data were collected. 

\begin{figure*}
\centering
	\subfigure[b][Left maneuver]{
	\includegraphics[width=2.9in, height=1.7in]{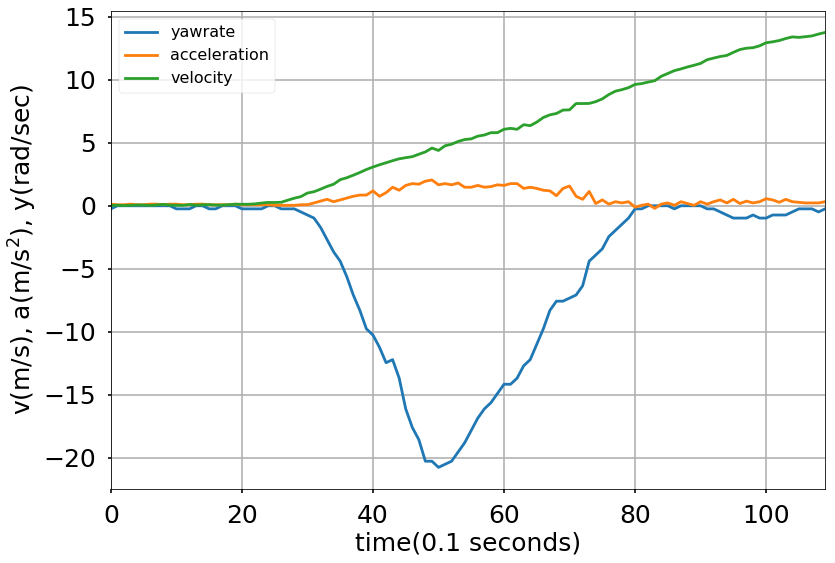}
	}
\subfigure[b][Right maneuver]{
	\includegraphics[width=2.9in,         height=1.7in]{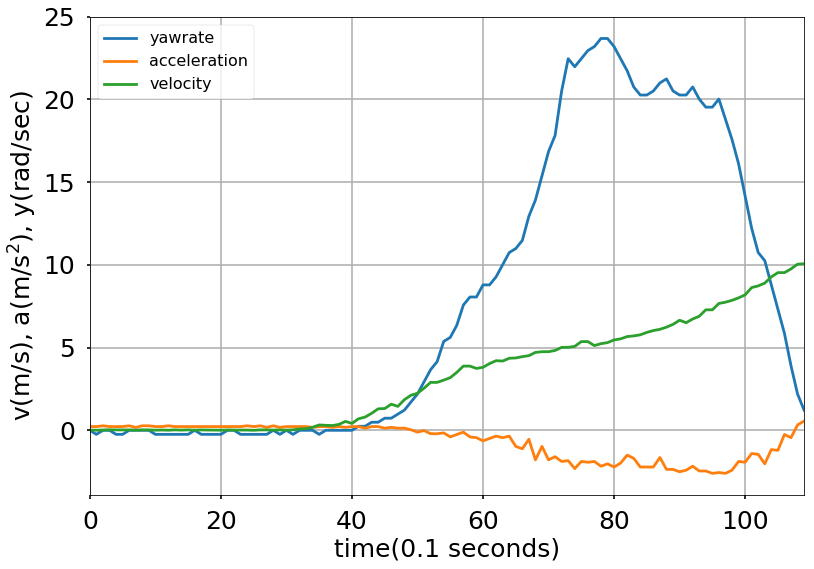}

	}
\subfigure[b][Straight maneuver]{
	\includegraphics[width=2.9in,         height=1.7in]{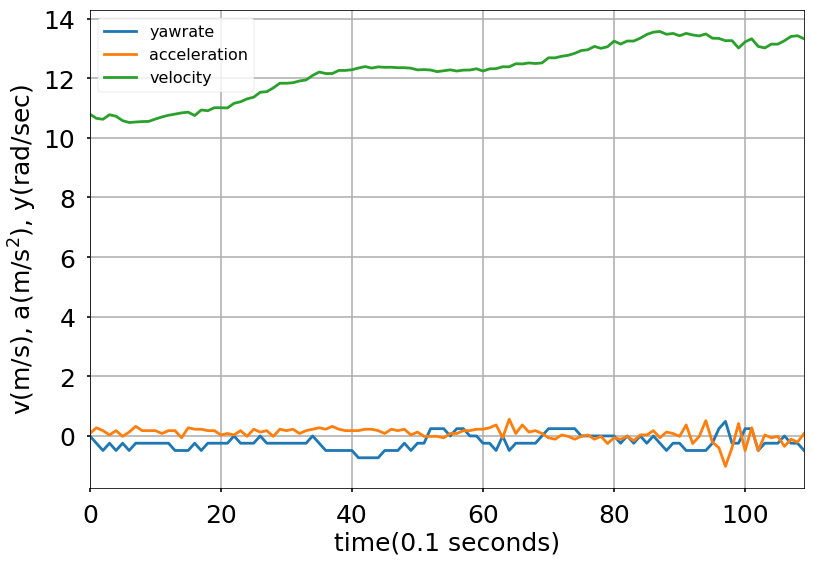}
	}
\subfigure[b][Stop maneuver]{
	\includegraphics[width=2.9in,         height=1.7in]{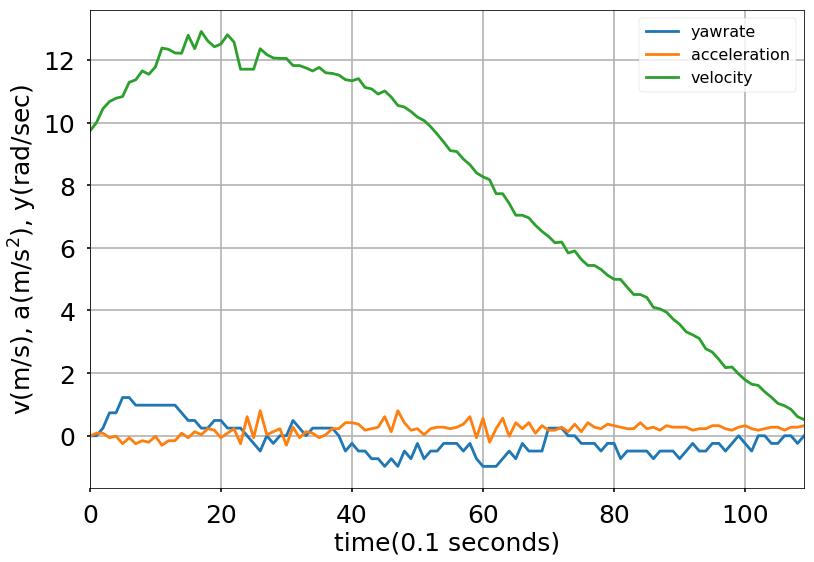}
	}
	\caption{Time series observation of velocity, yaw-rate and acceleration for 11 seconds time-span.}
	\label{fig:datavis}
\end{figure*}



A daily routine roads often used by the participant are selected for the experiment to represent ordinary road circumstances conditions. In this study, the scenario of interest is the road intersection where the driver can go straight, turn left, turn right, or stop. The timestamps that represent the ground truth scenarios are manually labeled based on the collected video data. For example, in the video, if the target vehicle come to the intersection and goes straight, that timespan the vehicle spent in the intersection vicinity is labeled as the "Straight" maneuver, and the corresponding ground truth data of the vehicle within that timespan get the "Straight" label. Figure \ref{fig:datavis} shows the ground truth time series data for ll seconds for a different example, maneuvers. As shown in \cite{amsalu2015driver, amsalu2016driver,amsalu2017driver, khosroshahi2016surround} these observation can capture and describe drivers behavior at road intersection . 

The labeled time-series data were used to train and test the proposed model for estimating the intention of the driver. In this work, the velocity and yaw-rate of the vehicle are considered to develop the model. Acceleration provides useful information about change in velocity.  However, as shown in Figure \ref{fig:datavis}, since the acceleration for the different driver's intentions is almost the same, it was not used as input to the model.

As we are modeling the problem as sequence to sequence task, the proposed LSTM model takes discretized input. Based on the ground truth which is observed from the recorded video, the selected input feature data is discritized into categories for each time-step as follows:
\begin{itemize}
    \item \textit{Velocity (V)}:- categorized into two classes; low velocity for V $< $ $10$ m/s (Class $0$) and high velocity for values V $\geq$ $10$ m/s (Class $1$). Accordingly, stop, left and right turn maneuvers are in low-velocity class.
    \item \textit{Yaw-rate ($Y$)}:- grouped into three classes; high negative turns for $Y$ $<$ $-3$ rad/s (Class $0$), medium for $-3$ rad/s $\leq$ $Y$ $\leq$ $3$ rad/s (Class $1$), and high positive turns for $Y$ $>$ $3$ rad/s (Class $2$). The $3$ rad/s and $-3$ rad/s thresholds are selected based on the recorded observation to represent the distinction between the positive right, negative left and small random turns.
\end{itemize}
  
The classes from velocity and yaw-rate are again combined and categorized into $6$ discrete type symbols that are $2$ velocity classes times $3$ yaw-rate classes. For example, class-$0$ velocity and class-$0$ yaw-rate are categorized into symbol-$1$, and  class-$0$ velocity and and class-$1$ yaw-rate are categorized under symbol-$2$, etc. Accordingly, the total combination of all the classes results overall $6$ type of symbols.
  
As shown in Figure \ref{fig:datavis}, each selected maneuver's, velocity and yaw-rate, contains 110 time-series data, which are recorded for a timespan of $11$ seconds ($5$ sec before and $6$ sec after the intersection). For real-time implementation, smart infrastructure or computer vision can be used to notify the vehicle when it is approaching to an intersection. These data is then categorized into a sequence of $110$ symbols based on the $6$ types of symbol as explained above. As shown in Figure \ref{fig: arch}, these discrete sequences of symbols are then used to train the proposed LSTM model.  

\section{RESULTS}
The prediction performance of the model is evaluated based on accuracy, recall and F1-score which are defined as follows:
\begin{itemize}
    \item Precision (P):
    \begin{equation}
    P = \dfrac{T_P}{T_P + F_P}.
    \end{equation}
    \item Recall (R):
    \begin{equation}
    R =  \dfrac{T_P}{T_P + F_N}.
    \end{equation}
    \item F1-score:
    \begin{equation}
    F1  = 2 \times \dfrac{P \times R}{P + R}.
    \end{equation}
\end{itemize}
Where $T_P$, $F_P$ and $F_N$ represent true positive, false positive and false negative prediction respectively. A confusion Matrix is also used which shows ground truth observation in the row and the predictions in the column direction.  In this study, the proposed model is implemented in Keras\footnote{https://keras.io/} with TensorFlow \footnote{https://www.tensorflow.org/} deep neural network framework and other conventional models used for comparison are implemented using scikit-learn \footnote{https://scikit-learn.org/}.

The data set contains in total $2970$ datapoints, where straight (990), stop (770), right (660) and left (550) datapoints at every time step $t=\tau$. The dataset is splitted into training ($70\%$) and test ($30\%$). The discretized and symbolized time series observations of the selected features (yaw-rate and velocity), are used to train the proposed model. As discussed in the previous section and shown in Figure \ref{fig: arch},  a sequence of the discrete symbols used as input to the model. The output of the model is sequences of symbols representing each of the maneuvers as S(Straight), P(Stop), R(Right Turn) and L(Left Turn). The length of the symbolized input sequence fed to the model is $T_{x} = 110$ and the corresponding predicted output sequence is $T_{y} = 110$. The size of the sequence ($110$) represents the number of datapoints recorded for $11$ seconds with 0.1 seconds time step during the maneuver that took place at the intersection. The proposed model predicts the trajectory of the vehicle for a given sequence of input symbols at each time step. 

\begin{figure}[!h]
    \centering
    \includegraphics[width=3.2in,         height=1.5in]{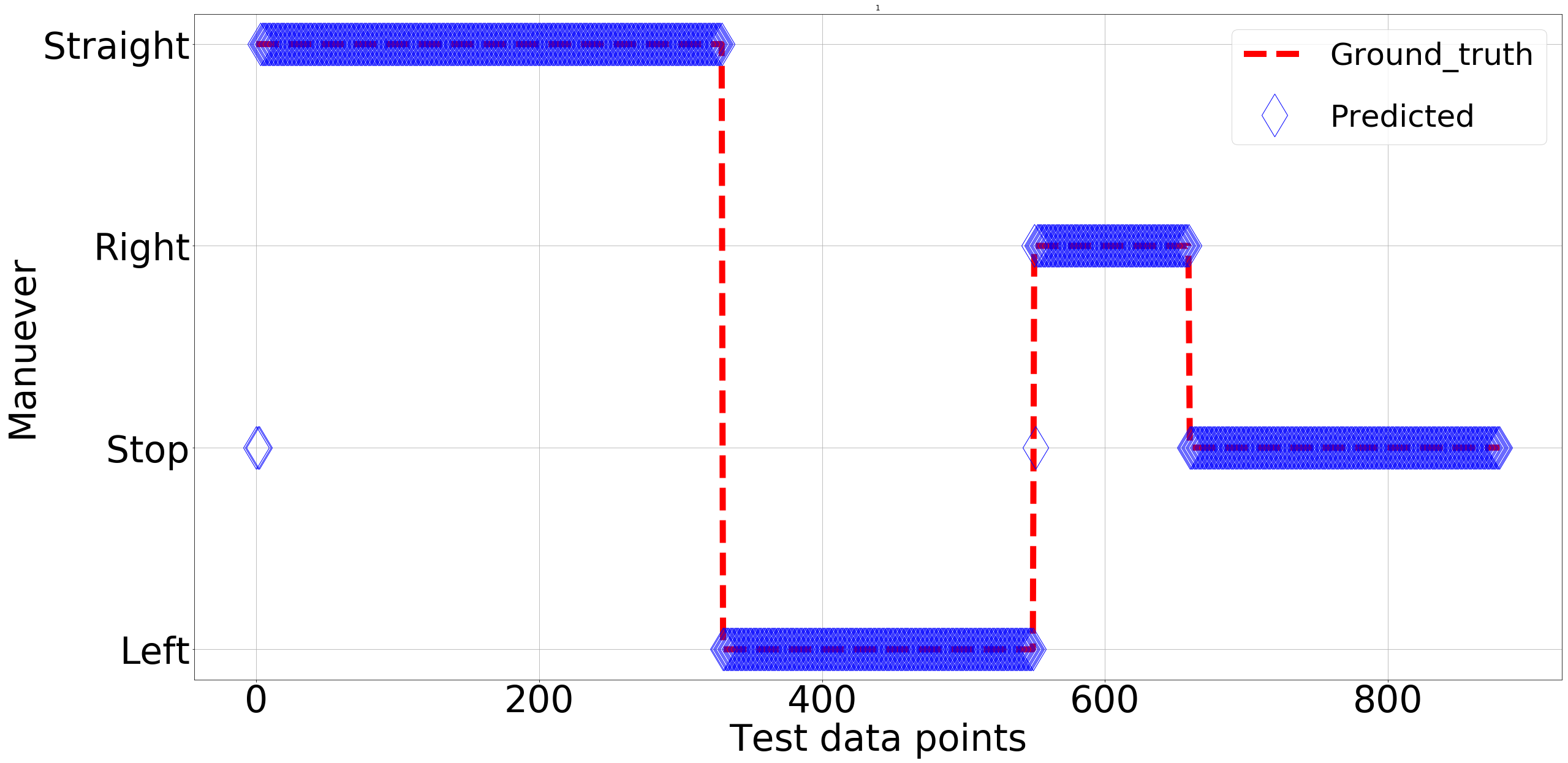}
    \caption{The ground truth and predicted values plot of the whole test dataset.}
    \label{fig:accuracy}
\end{figure}

\begin{table}[]
\centering
\caption{Proposed model performance using 30\% testing set maneuvers as presented in confusion matrix.}
\label{my-label}
\begin{tabular}{|c|c|c|c|c|}
\hline
\multirow{2}{*}{\begin{tabular}{c} \textbf{Ground-truth} \\ \textbf{Maneuvers} \\ \textbf{}\end{tabular}} & \multicolumn{4}{c|}{\textbf{Predicted Maneuvers }} \\ \cline{2-5}
                & \textit{Straight}  &\begin{tabular}{c} \textit{Left} \\ \textit{Turn}\end{tabular} & \begin{tabular}{c} \textit{Right} \\ \textit{Turn}\end{tabular}  &  \textit{Stop}    \\ \hline
                           Straight (S)    & 218 & 0   & 0   & 2   \\ \hline
                           Left Turn (L)  & 0   & 219 & 0   & 1   \\ \hline
                           Right Turn (R)  & 0   & 0   & 110 &    \\ \hline
                                  Stop (P)& 0   &  0  & 0	 & 330    \\ \hline
\end{tabular}
\\
Accuracy = 99.65 $\%$

\label{tab:conf}
\end{table}
The trained model performance is tested on unseen 30\% of dataset and the result is shown in a confusion matrix on Table \ref{tab:conf}. The model achieved 99.65 $\%$ accuracy and 99.64 $\%$ recall. Based on the discretized input sequence of observations the model predicts the corresponding output sequence of vehicle maneuver as; straight (S), right turn (R), left turn (L) and stop (P), as shown in Figure \ref{fig:accuracy}. All the maneuvers shown are from the testing set of the dataset.

\begin{figure}[]
    \centering
    \includegraphics[width=3.3in,         height=2.5in]{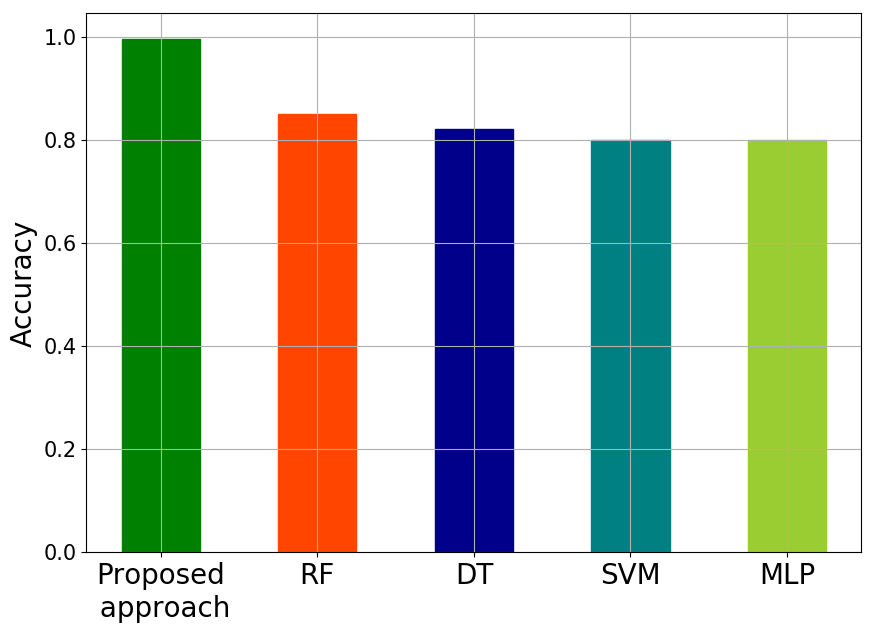}
    \caption{Proposed model performance comparison against conventional machine learning models, where RF(random forest), DT(decision tree), SVM(support vector machine) and MLP(multilayer perceptron).}
    \label{fig:modelComparession}
    \end{figure}

The proposed model is compared with other four widely used conventional machine learning algorithms, Decision Tree (DT), Multilayer Preceptor (MLP), Random Forest (RF) and Support Vector Machine (SVM). As shown in the Figure \ref{fig:modelComparession}, the proposed approach significantly outperformed other conventional techniques. This is largely is due conventional machine learning algorithms don't have inherent ability to extract temporal information out of time-series data. As the driving data is time-series, convectional algorithms performed poor compared to the proposed technique that has inherent ability of capturing temporal dependencies.

\begin{figure}[!h]
    \centering
    \includegraphics[width=3.1in,         height=1.2in]{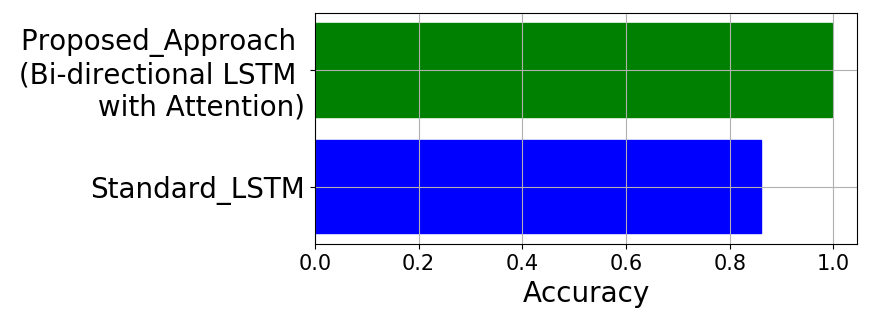}
    \caption{Proposed model performance Vs standard-LSTM.}
    \label{fig:ProposedVsLSTM}
    \end{figure}
Furthermore, we compared the proposed bidirectional LSTM with attention mechanism model against a standard LSTM model to see the advantage of using bidirectional and attention mechanism. As shown in Figure \ref{fig:ProposedVsLSTM}, the proposed model outperformed the standard LSTM-model. As demonstrated in \cite{vaswani2017attention}, attention mechanism allows the LSTM network to learn where to focus on input sequence that enables more effective learning ability than standard-LSTM. Besides, as compared in \cite{graves2005bidirectional}, bidirectional LSTM enables the network to exploit more temporal information by running inputs from preceding and succeeding part of the input sequence. Thus, these two techniques enable the proposed approach to achieve much greater result than its counterpart standard-LSTM.

\begin{table}[!h]
	\caption{Model comparison}
	\centering
	\begin{tabular}{|c|c|c|c|}
		\hline
		Model & Accuracy & Recall & F1-score \\ \hline\hline
		Support Vector Machine & 0.8 & 0.74 & 0.74 \\\hline
		Multi-layer Perceptron & 0.8 & 0.79 & 0.79\\ \hline
		Decision Tree & 0.82 & 0.81 & 0.80 \\\hline
		Random Forest & 0.85 & 0.85 & 0.85 \\\hline
		Standard LSTM & 0.86 & 0.82 & 0.77 \\ \hline \hline
		Proposed Approach & 0.9965 & 0.9964 & 0.9964 \\ \hline

	\end{tabular}
	\label{tab:summary}
\end{table}   
Generally, as summarized in Table \ref{tab:summary}, the proposed technique has addressed the issue of estimating drivers' intention near a road intersection and out-performed other techniques. 

\section{CONCLUSION}
In this paper, bidirectional LSTM with attention mechanism is used to address the problem of driver's behavior estimation near road intersection. A real-world time-series vehicular data is used in this study. The HSS framework combined with proposed LSTM model is used to mathematically relate the continuous vehicle dynamics data to the driver's discrete decisions. The yaw-rate and velocity data of a vehicle approaching intersection is recorded and discretized into a sequence of symbols at each time step and used to train and test the proposed model. The model estimates the driver's maneuver intention as Straight, Stop, Right turn and Left turn. Result shows that the model predicts with a high accuracy and outperformed other approaches substantially. In future work, the proposed method can be applied in estimating drivers intention in high-way merging, lane changing and other near-crash events.






\section*{ACKNOWLEDGMENT}
This work is based on research supported by NASA Langley Research Center under agreement number C16-2B00-NCAT, and Air Force Research Laboratory and Office of the Secretary of Defense (OSD) under agreement number FA8750-15-2-0116.

\bibliographystyle{IEEEtran}
\bibliography{root}

%
%

\end{document}